\RequirePackage{fix-cm}
\RequirePackage{amsmath}
\documentclass[twocolumn]{svjour3}          % twocolumn
\smartqed  % flush right qed marks, e.g. at end of proof
\usepackage{graphicx}
\usepackage{array}
\usepackage{multirow}
\usepackage{rotating}
\usepackage{pifont}
\usepackage{natbib}
\usepackage[citecolor=blue, colorlinks=true, citebordercolor={1 1 1}]{hyperref}
\usepackage{algorithm}
\usepackage{algorithmicx}
\usepackage{algpseudocode}
\usepackage{lipsum}
\usepackage{mathtools}
\usepackage{cuted}
\usepackage{breakcites}
\usepackage{microtype}
\usepackage{chngcntr}
\usepackage{soul}

\newcommand{\cmark}{\ding{51}}
\newcommand{\xmark}{\ding{55}}
\begin{document}

\title{Early-exit Convolutional Neural Networks
%\thanks{Grants or other notes
%about the article that should go on the front page should be
%placed here. General acknowledgments should be placed at the end of the article.}
}
%\subtitle{Do you have a subtitle?\\ If so, write it here}

%\titlerunning{Short form of title}        % if too long for running head

\author{Edanur Demir         \and
        Emre Akbas
}

%\authorrunning{Short form of author list} % if too long for running head

\institute{E. Demir \at
Middle East Technical University, Ankara, Turkey \\
\email{e181920@metu.edu.tr}
\and
E. Akbas (corresponding author) \at
Middle East Technical University, Ankara, Turkey \\
Tel.: +90-312-210 5522\\
Fax: +90-312-210 5544\\
 \email{emre@ceng.metu.edu.tr}\\
Web: \url{http://user.ceng.metu.edu.tr/~emre/}
}

\date{Received: date / Accepted: date}
% The correct dates will be entered by the editor

\maketitle
\begin{abstract}
This paper is aimed at developing a method that reduces the computational cost of convolutional neural networks (CNN) during inference. Conventionally, the input data pass through a fixed neural network architecture. However, easy examples can be classified at early stages of processing and conventional networks do not take this into account. In this paper, we introduce 'Early-exit CNNs', \textit{EENets} for short, which adapt their computational cost based on the input by stopping the inference process at certain exit locations. In EENets, there are a number of exit blocks each of which consists of a confidence branch and a softmax branch. The confidence branch computes the confidence score of exiting (i.e. stopping the inference process) at that location; while the softmax branch outputs a classification probability vector. Both branches are learnable and their parameters are separate. During training of EENets, in addition to the classical classification loss, the computational cost of inference is taken into account as well. As a result, the network adapts its many confidence branches to the inputs so that less computation is spent for easy examples. Inference works as in conventional feed-forward networks, however, when the output of a confidence branch is larger than a certain threshold, the inference stops for that specific example. The idea of EENets is applicable to available CNN architectures such as ResNets. Through comprehensive experiments on MNIST, SVHN, CIFAR10 and Tiny-ImageNet datasets, we show that early-exit (EE) ResNets achieve similar accuracy with their non-EE versions while reducing the computational cost to 20\% of the original. Code is available at \href{url}{https://github.com/eksuas/eenets.pytorch} 
% \PACS{PACS code1 \and PACS code2 \and more}
% \subclass{MSC code1 \and MSC code2 \and more}
\end{abstract}

\section{Introduction}
\label{section:b1}

\begin{figure}
\centering
\includegraphics[width=.45\textwidth]{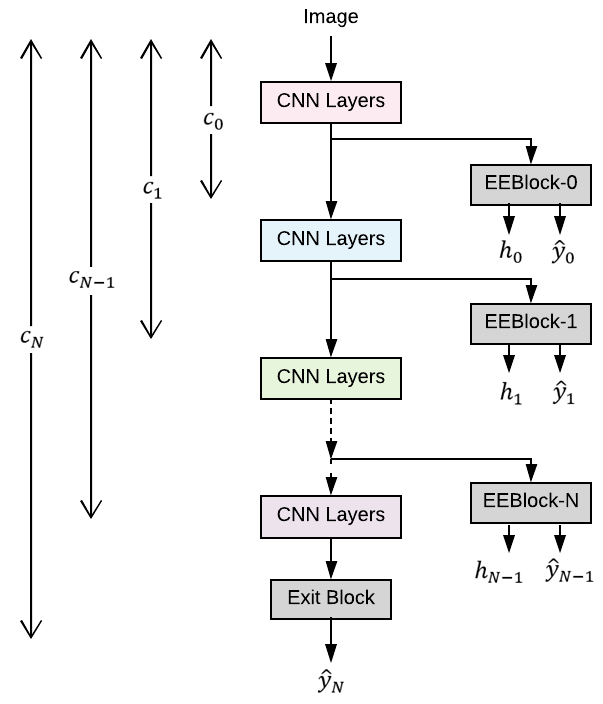}
\caption[Architectural overview of EENets]{Architectural overview of EENets. An early-exit block (shown with gray color) can be added at any location. If, at a certain early-exit block, say the $i^{th}$ one, the network is sufficiently confident (i.e. $h_i>0.5$), then the execution is terminated at that point and the network's output is set to $\hat{y}_i$. $c_i$ denotes the computational cost (in terms of the total number of floating-point operations) upto the $i^{th}$ early-exit block. ``CNN layers'' are classical computation blocks that may be composed of one or more convolutional or fully-connected layers and non-linear activation functions. EENets aim to strike a balance between minimizing the computational cost and maximizing the accuracy.
}
\label{fig:overview}
\end{figure}

Deep neural networks are power-hungry. They typically need powerful processing units (i.e. GPU cards) in order to run in a reasonable amount of time. Reducing their computational cost with minimal degradation in accuracy is an important goal that has been approached from several different directions. One promising way to this end is to make the network adapt its computational cost to the input during inference. This idea has recently been explored in many ways. Researches have proposed early termination networks \citep{msdnet,branchynet,sacrificing,sact,cdl}, layer skipping networks \citep{adanet,skipnet,blockdrop}, specialized branches with wide networks \citep{hydranet}, adaptive neural trees \citep{ant}, cascaded networks \citep{ann} and pruning methods such as channel gating networks \citep{channelgate}.

Conventionally, the input data pass through a fixed neural network architecture. However, easy examples can be classified at early stages of processing and conventional networks do not take this account. In order to reduce the computational cost, the methods mentioned above aim to adapt the computation graph of the network to the characteristics of the input instead of running the fixed model that is agnostic to the input. Our work in this paper can be categorized under the ``early termination networks'' category.

\begin{sloppypar}
In this paper, we introduce ``Early-exit CNNs'', \textit{EENets} for short, which adapt their computational cost based on the input itself by stopping the inference process at certain exit locations. Therefore, an input does not have to flow through all the layers of the fixed network; on the contrary, the computational cost can be significantly decreased based on the characteristics of inputs. Figure \ref{fig:overview} shows the architectural overview of the Early-exit Convolutional Neural Networks (EENets).
\end{sloppypar}

In EENets, there are a number of exit blocks each of which consists of a confidence branch and a softmax classification branch. The confidence branch computes the confidence score of exiting (i.e. stopping the inference process) at that location; while the softmax branch outputs a classification probability vector. Both branches are trainable and they are independent of each other. These exit blocks constitute a very small part of the overall network (e.g. a single exit blocks constitutes $\sim$0.0002\% of the parameters in a EENet-110 (akin to ResNet-110) designed for a 10-class dataset). In short, the additional parameters coming from early-exit blocks do not significantly increase the computational cost, hence, they can be ignored.

During training of EENets, in addition to the classical classification loss, the computational cost of inference is taken into account as well. As a result, the network adapts its confidence branches to the inputs so that less computational resources are spent for easy examples. That is, when the confidence (i.e. $h_i$ in Fig. \ref{fig:overview}) at a certain exit block is larger than a predetermined threshold, the inference process stops for that specific example. 

Defining a proper way to train a EENet network is important because the model could be biased towards wrong decisions such as early or late termination. These wrong decisions could either decrease accuracy or increase the computational cost unnecessarily. Deciding at which point an input can be classified and the execution can be terminated is the key challenge of the problem. To address this problem, we propose a novel loss function which balances computational cost and classification loss in a single expression enabling the training of the base neural network and all the exit blocks simultaneously. 

Our experiments show that EENets achieve similar accuracy compared to their counterpart ResNets \citep{resnet} with relative computational costs of $30\%$ on SVHN \citep{svhn}, $20\%$ on CIFAR10 \citep{cifar} and $42\%$ on Tiny-ImageNet \citep{tiny-imagenet} datasets. 
%\hl{In the models which have a large capacity such as EENet-152, the cost savings can be upto 50x.} 
%In such deep models, EENets not only maintain but also improve the accuracy a little probably due to the regularizing effect of using less parameters.

% Contributions
\subsection{Contributions}
In the context of  previous related work, our  contributions with the introduction of EENets are as follows:
\begin{itemize}
\item EENet has a single stage training as opposed similar previous work which are trained in multiple stages.
\item EENets are compact models not requiring additional hyper parameters such as non-termination penalty or confidence threshold variables.
\item The confidence scores of EENets are learnable and they do not depend on  heuristic calculations. As a consequence, their initialization is not an issue and they can be initialized just like other parameters of the network.
\item Our loss function considers both accuracy and cost simultaneously and provides a trade-off between them via an hyper-parameter.
\item All exit blocks of an EENet are fed by all inputs even if some inputs are classified in early stages of the model. This avoids a possible dead unit problem (which is a frequent problem in previous work) where some layers are not trained at all.
\end{itemize}

\section{Related Work}
\label{section:b2}

Neural networks that adapt their computations based on the input’s characteristic can be examined in the following main categories: early termination networks, layer skipping networks, specialized branches with wide networks, neural trees, cascaded networks and pruning methods such as channel gating networks. In addition, some studies focus on the confidence degree of image classification \citep{confnet}.

\subsection{Early Termination Networks}
Early termination network are based on the idea that it might not be necessary to run the whole network for some inputs. Similar to EENets, early termination networks \citep{msdnet,branchynet,sacrificing,sact,cdl} have multiple exit blocks that allow early termination based on an input’s characteristics. All of these studies have some kind of confidence scores to decide early termination.

One of the early termination networks, BranchyNets \citep{branchynet} have multiple exit blocks each of which consists of a few convolutional layers followed by a classifier with softmax. In other words, BranchyNets have one head just for classification at their exit blocks. The exit blocks of  \citet{sacrificing}'s model are composed of pooling, two fully-connected (FCs) and batch normalization layers. Like BranchyNets, one conventional head at an exit block is trained for classification. The confidence scores are derived via some heuristics. In the training procedure of the model of Beretizshevsky and Even, the weights of only convolutional and the last FC layers are firstly optimized. Later, the remaining FC layers are optimized, one by one. On the other hand, MSDNets \citep{msdnet} have multi-scaled features with dense connectivity among them. Exits of MSDNets consist of two convolutional layers followed by a pooling and a linear layer. However, similar to BranchyNets, MSDNets do not have confidence branches at their exit blocks.

In these models, the confidence scores are derived from the predicted classification results (i.e. the maximum over the softmax). Because such confidence scores are not learnable, deciding on the termination criteria or threshold of an exit branch becomes an important issue. The exit threshold providing the maximum accuracy should be empirically discovered in these models. Unlike EENets, the loss functions of these models do not encourage an early-exit by considering the computational cost. In addition, they have a \emph{dead layer problem} coming from improper initialization of the confidence scores in the training. The scores may be biased to exit always early and deeper layers may not receive learning signals properly. To avoid this situation, the model of \citet{sacrificing} use a multi-stage training. In the first stage, it optimizes all the convolutional weights together with the weights of the last FC layers. After that, it optimizes the weights of the remaining FC components, one by one.

Spatially Adaptive Computation Time for Residual Networks, shortly SACTs \citep{sact}, is another study in the category. Exit blocks of the model consist of a pooling and a fully-connected layer followed by a sigmoid function like our model. However, the final confidence score of early termination (namely ``halting score" in their paper) is calculated by the cumulative learnable scores of the previous exit blocks. As soon as the cumulative halting score reaches a constant threshold (i.e. $T \geq 1.0$), the computation is terminated. Unlike EENets, the classification output vector of SACTs (i.e. the output of the softmax branch) is derived from weighted summation of the inputs of the confidence branches so far. While EENets directly train the confidence scores by taking them into account in the loss function, SACTs employ the number of executed layers as non-termination penalty in the loss function. Another work, Conditional Deep Learning (CDL) \citep{cdl} has multiple exit blocks each of which consists of just a linear classifier. Starting from the first layer, linear classifiers are added to the end of each convolutional layer iteratively as long as this addition process does not decrease the accuracy. In CDL, a user defined threshold is used to decide if the model is sufficiently confident to exit. The training procedures of SACTs and CDLs are also multi-stage.

\subsection{Layer Skipping Networks}
Layer skipping networks \citep{adanet,skipnet,blockdrop} adapt their computation to the input by selectively skipping layers. In these networks, a gating mechanism determines, for a specific input, whether the execution of the layer  can be skipped. The main challenge here is learning the discrete decisions of the gates. AdaNets \citep{adanet} use Gumbel Sampling \citep{gumbel} while SkipNets \citep{skipnet} and BlockDrop \citep{blockdrop} apply reinforcement learning to this end. None of these models has a separate confidence branch at the gate blocks. 
Similar to the early-exit blocks of early termination nets, the gates of the layer skipping networks may die and lose their functionality if they incline to be too much turned off during training. Thus, the actual capacity usage decreases. On the other hand, if the gates tend to be turned on, the networks cannot reduce computational cost effectively. As a result, the networks can not only perform as counterpart static models but also spend additional computational cost for the gate functions (i.e. the same capacity with more cost). In order to avoid such cases, the gate blocks require to be initialized carefully and trained properly. Thus, models in this category (i.e. layer skipping networks) have a complicated multi-stage training.

\subsection{Specialized Branches with Wide Networks}
As wide networks, HydraNets \citep{hydranet} is another approach in the area. HydraNets contain distinct branches specialized in visually similar classes. HydraNets possess a single gate and a combiner. The gate decides which branches to be executed at inference. And the combiner aggregates features from multiple branches to make a final prediction. In training, given a subtask partitioning (i.e. dividing dataset into visually similar classes), the gate and the combiner of the HydraNets are trained jointly. The branches are indirectly supervised by the classification predictions after combining the features computed by the top-$k$ branches.

\subsection{Neural Trees}
Adaptive Neural Trees, ANTs \citep{ant}, can be considered as a combination of decision trees (DTs) with deep neural networks (DNNs). It includes the features of the conditional computation of DTs with the hierarchical representation learning and gradient descent optimization of DNNs. ANTs learn routing functions of a decision tree thanks to the training feature of DNNs. While doing this, instead of a classical entropy, ANTs use stochastic routing, where the binary decision is sampled from Bernoulli distribution with mean $r^{\theta}(x)$ for input $x$ ($r^{\theta}$ can be a small CNN). However, ANTs are trained in two stages: \textit{growth phase} during which the model is trained based on local optimization and \textit{refinement phase} which further tunes the parameters of the model based on global optimization. %The training process of ANTs is complicated because of the refinement phase.

\subsection{Cascaded Networks}
Some other approaches focus on cascaded systems. The model by \citet{ann} adaptively chooses a deep network among the-state-of-arts such as AlexNet \citep{alex}, GoogleNet \citep{googlenet}, and ResNet \citep{resnet} to be executed per example. Each convolutional layer is followed by the decision function to choose a network. But it is hard to decide if termination should be performed just by considering a convolutional layer without employing any classifier. It has a multi-stage training procedure where the gates are trained independently from the rest of the model.  

\subsection{Pruning Methods}
Channel Gating Neural Networks \citep{channelgate} dynamically prune computation on a subset of input channels. Based on the first $p$ channels, a gate decides whether to mask the rest of the channels. Similar to SACTs \citep{sact}, when the classification confidence score reaches a threshold, the remaining channels are not computed.

\begin{table*}
\centering
\begin{tabular}{|l||c|c|c|c|}
\hline
\multirow{2}{*}{\bf Model} & \bf Single Stage & \bf Non-specialized & \bf Learnable & \bf Loss Func. \\
& \bf Training & \bf Initialization & \bf Confidence & \bf Includes \\
\hline
\hline
AdaNet \citep{adanet} 
& \cmark & \xmark & \cmark & acc and \#exec. layers  \\
ANT \citep{ant}  
& \xmark & \xmark & -      & accuracy                 \\
\citet{sacrificing}  
& \xmark & \xmark & \xmark & accuracy                 \\
BlockDrop \citep{blockdrop}  
& \xmark & \xmark & \cmark & accuracy and cost        \\
\citet{ann}  
& \xmark & \cmark & \cmark & accuracy and cost        \\
BranchyNet \citep{branchynet}  
& \cmark & \xmark & \xmark & accuracy                 \\
CDL \citep{cdl}   
& \xmark & \xmark & \xmark & accuracy and cost        \\
Channel gating \citep{channelgate}   
& \xmark & \xmark & \cmark & accuracy and cost        \\
HydraNet \citep{hydranet}  
& \cmark & \xmark & -      & acc of top-k branches    \\
MSDNet \citep{msdnet}  
& \cmark & \cmark & \xmark & acc of top-k classifier  \\
SkipNet \citep{skipnet}  
& \xmark & \xmark & \xmark & accuracy and cost        \\
SACT \citep{sact}  
& \xmark & \xmark & \cmark & acc and \#exec. layers  \\
EENet (Ours) & \cmark & \cmark & \cmark & accuracy and cost \\
\hline
\multicolumn{5}{c}{}
\end{tabular}
\caption[Differences with related work]{\textbf{Differences with related work.} The features of the related work are compared in terms of whether they have a single stage training, a non-specialized initialization process and learnable confidence scores in the table above. Check mark represents whether the model has the feature or not. The last column shows what their loss functions include (e.g. the classical classification loss as the accuracy or the number of executed layers (\# exec. layers) as the computational cost). The term of “accuracy and cost” just shows that the loss function takes both of them into account but note that the accuracy and cost values of different models can be obtained in different ways. Some features may not be applicable for some models. In such cases, we use “-” symbol.}
\label{table:related_work}
\end{table*}

\subsection{Novelties of EENets}
Table \ref{table:related_work} summarizes the differences between our proposed EENets and related previous work. 

As discussed above, many models from different categories have the \emph{dead layer/unit problem}. In EENets, we avoid this problem with our novel loss function (described in Section \ref{section:b3}) which enables the training of all exit blocks by all inputs, even if some inputs are classified in the early stages of the model. 

Another contribution of EENets is the separate confidence branches at their exit blocks. Unlike most of the previous adaptive computational approaches, the confidence scores of EENets are trainable and do not depend on  heuristic calculations. 
Having separate learnable parameters allows the confidence branches to be not biased towards classification results. Their initialization is not an issue and they can be initialized just like other parameters of the network. This separate confidence branches approach makes EENets easier to use/train compared to the previous work. 

Another novelty of EENets is the loss function that takes both accuracy and the cost spending into account simultaneously and provides a trade-off between them through the confidence scores. In contrast to most of the previous studies, our cost values employed in the loss function are not hyper-parameters but are based on the  actual number of floating-point operations. Unlike most of the previous studies, EENets have a single stage training in spite of having multiple exit-blocks. EENets do not require additional hyper-parameters such as non-termination penalty or confidence threshold variables.

\section{Model}
\label{section:b3}

In this section, we describe the architecture of EENets, the details about types of exit blocks, how to distribute early-exit blocks to a network, feed-forward and backward phases of the model and the proposed loss function.

\subsection{Architecture}
Any given convolutional neural network (CNN) can be converted to an early-exit network by  adding early-exit blocks at desired locations. To achieve this, first, one has to decide how many exit-blocks are going to be used. This is a design choice. Next, the locations where to connect the exit-blocks need to be decided. We propose various ways of doing this in Section \ref{subsection:distributing-blocks}. Finally, one needs to decide which type of exit-blocks to use. In the following paragraphs, we describe three different types of exit-blocks. 

%In this paper, we explain EENets through the ResNet architectures \citep{resnet} since they are widely used and they yield state-of-the-art results in many problems. 

%Because we explain EENets through ResNets, the main flow of EENets can be considered being composed of the identity skip-connections and the layer functions such as convolution, pooling etc. Formally, consider $F_l(\cdot)$ and $\mathbf{x}_l$ as the function of the $l^{th}$ layer and the output of this function, respectively. The main flow can be defined as:

% \begin{equation}
% \qquad \qquad \qquad \quad
% \mathbf{x}_l=\mathbf{x}_{l-1} + F_l(\mathbf{x}_{l-1})
% \label{eq:resnet}
% \end{equation}

%Beside the main network, the architecture of EENets has multiple early-exit (EE) blocks. 
Each EE-block consists of two fully-connected (FC) heads, namely the confidence branch and  the classification softmax branch. Both take the same channel-based feature maps (from the previous layer) as input.

\begin{figure}
\centering
\includegraphics[width=.48\textwidth]{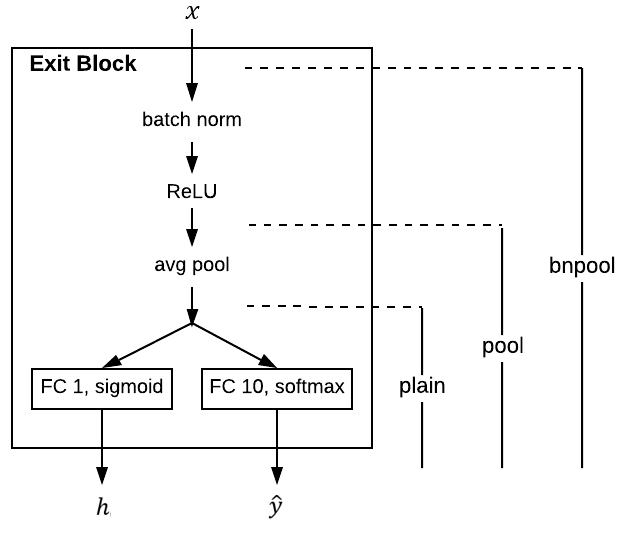}
\caption[Architecture of \textit{Plain}, \textit{Pool}, and \textit{Bnpool} early-exit blocks]{Architecture of \textit{plain}, \textit{Pool}, and \textit{Bnpool} early-exit blocks. The \textit{plain}-type exits are composed of just separate fully-connected (FC) layers and input of that block is directly processed in the FC branches. The \textit{Pool} exits have a global average pooling layer before FC branches. Lastly, the \textit{Bnpool}-type exit blocks consist of a batch normalization layer followed by a ReLU activation and a global average pooling layer. The input of the early-exit block, $x$, passes onto these layers before entering the separate FC branches. $h$ and $\hat{y}_{n}$ denote the confidence score and the predicted classification label. “$\mathrm{fc}$ $X, \mathrm{activation}$” denotes the fully-connected heads which have $X$ number of outputs. The $\mathrm{activation}$ is the last activation function of branches.}
\label{fig:blockTypes}
\end{figure}

We define three types of early-exit blocks, namely, \textit{Plain}, \textit{Pool} and \textit{Bnpool}. The \textit{Plain}-type exit is composed of two separate fully-connected (FC) layers and input feature maps  are directly fed to these FC branches. The \textit{Pool} exit has a global average pooling layer before the FC branches. Lastly, the \textit{Bnpool}-type exit block consists of a batch normalization layer followed by a ReLU activation and a global average pooling layer before the FC confidence and classification branches. Figure \ref{fig:blockTypes} presents the architectures of the three different types of early-exit blocks.

In the \textit{Pool} and \textit{Bnpool} early-exit blocks, the size of the input feature map is reduced by global average pooling that is denoted by $z(\mathbf{x})$. The purpose of this is to reduce the computational cost at early-exit blocks. 
%Consequently, they can decide to terminate the execution, in a shorter amount of time. 
Our experiments show that the early-exit blocks that have a global average pooling layer yield more accurate results  (Section \ref{section:b4}). 
%The following equations are constructed on the idea of \textit{Pool} early-exit blocks. 
The average pooling function is as follows:

\begin{equation}
z_{n,c}(\mathbf{x}) = \frac{1}{H*W} \sum_{i=1}^{H} \sum_{j=1}^{W} \mathbf{x}_{n,c,i,j}
\label{eq:avg_pool}
\end{equation}

\noindent where $n$ denotes the batch size and $c$ denotes the number of channels. $H$ and $W$ denote height and width of the feature maps, respectively. 

\begin{sloppypar}
The pooled data passes onto two separate FC branches, the classification branch and the confidence branch. The number of outputs of the classification  branch is same as the number of classes in the dataset. This branch has a softmax activation at the end. The  confidence branch uses a sigmoid activation function which outputs a scalar  representing the confidence of the work at that specific exit block. %Both branches (or output heads) feed the network jointly and are back-propagated in the training.
\end{sloppypar}

Formally, let $\mathbf{x}$ be the input to the $n^{th}$ early-exit block. $\mathbf{x}$ is actually the output of the CNN layers (see Figure \ref{fig:overview}) immediately preceding the $n^{th}$ early-exit (EE) block. In the EE-block, two things are computed: (i) $\mathbf{\hat{y}}_{n}$, the class prediction vector, and (ii) $h_{n}$, the confidence level of the network for the prediction $\mathbf{\hat{y}}_{n}$. They are given in Eq. \eqref{eq:exit_block} where $\mathbf{w}_1$ and $\mathbf{w}_2$ are the parameters of separate fully-connected layers of the softmax and confidence branches, respectively.

\begin{equation}
\begin{aligned}
& \mathbf{\hat{y}}_{n} = \mathrm{softmax} (\mathbf{w}_1^{T} z(\mathbf{x})) \\
& h_{n} = \sigma (\mathbf{w}_2^{T} z(\mathbf{x}))
\end{aligned}
\label{eq:exit_block}
\end{equation}

%\begin{sloppypar}
%Note that the classification and the confidence branches have softmax and sigmoid activation functions, respectively. As a traditional multi-classification head, the classification branch of the early-exit blocks employs the softmax activation function. On the other hand, the confidence branch of EE-blocks uses the sigmoid activation function because it is expected to be estimate the probability of prediction correctness. The sigmoid function maps the output of the branch onto the interval $[0,1]$. The confidence score of the last exit layer (note that it is not an early-exit block) is set to 1 in order to guarantee to terminate the execution at the end of the model.
%\end{sloppypar}

%If the exit block is confident for classification (or termination), the data is classified, and the execution is terminated in inference phase. Otherwise, the data continues to pass through the remaining network.

%\begin{equation}
%\mathbf{x}_{l} = \begin{cases} E^{\hat{y}}(\mathbf{x}_{l-1}) & \text{if } E^{h} (\mathbf{x}_{l-1}) \geq T \\ 
%\mathbf{x}_{l-1}+F_l(\mathbf{x}_{l-1}) & \text{otherwise} \end{cases}
%\label{eq:feed_forward}
%\end{equation}

%In Equation (\ref{eq:feed_forward}), the main flow in a stage of the model is given. $\mathbf{x}_{l}$ denotes output of the $l^{th}$ layer. $T$ symbolizes the confidence threshold being set to $0.5$ which can be considered as the midpoint of decision mechanism because the sigmoid function $\sigma_{\mathrm{sigmoid}}(\mathbf{x}) \in [0,1]$ is used as an activation function of the confidence branch.

\subsection{Inference}
\label{subsection:inference}
The Early-exit Convolutional Neural Networks (EENets) have a certain threshold in order to decide early termination in the  inference procedure. If the confidence score of an early-exit block is above the threshold, the classification results of the current stage will be the final prediction. Each input is classified based on their individual confidence scores predicted by the early-exit blocks. Thus, one input can be classified and terminated early while others continue being processed by the model.

\begin{sloppypar}
Early termination threshold is $T = 0.5$. It is the midpoint of the output range of the sigmoid function (used by the confidence branches). The threshold is employed only in the inference phase. During training, all examples are processed by the entire network; thus, all early-exit blocks contribute to the loss function (see Section \ref{subsection:training}) for all examples even if some of them can be classified early. %Therefore, the threshold does not affect the training process. 
\end{sloppypar}

\begin{sloppypar}
The pseudo-code of the inference procedure of EENets is given in Algorithm \ref{alg:inference} where  $\mathrm{EEBlock}_{i}$ represents the $i^{th}$ early-exit (EE) block of the model and $\mathrm{CNN\_Layers}_{i}$ denotes the sequence of intermediate blocks (CNN layers) between ${(i-1)}^{th}$ EE-block and $i^{th}$ EE-block.  $\mathrm{CNN\_Layers}_{0}$ is the initial CNN layers of the model before entering any EE-block. $N$ denotes the total number of early-exit blocks. $h_{i}$ and $\hat{y}_{i}$ shows the confidence score and classification output vector of $i^{th}$ EE-block.
\end{sloppypar}

\begin{algorithm}
\caption{Inference of Early-exit Convolutional Neural Networks}
\begin{algorithmic} [1]
\State $ i \gets 0 $
\While {$ i < N $}
\State $ x \gets \mathrm{CNN\_Layers}_{i}(x) $
\State $ h_{i}, \hat{y}_{i} \gets \mathrm{EE\_Block}_{i}(x)$
\If {$ h_{i} \geq T $}
\State $ \textbf{return}$ $\hat{y}_{i}$
\EndIf 
\State $ i \gets i+1 $
\EndWhile
\State $ x \gets \mathrm{CNN\_Layers}_{i}(x) $
\State $ \hat{y} \gets \mathrm{Exit\_Block}(x)$
\State $ \textbf{return}$ $\hat{y} $
\end{algorithmic}
\label{alg:inference}
\end{algorithm}

\subsection{Training}
\label{subsection:training}
During training, the goal is to learn the parameters of the CNN and all the early-exit blocks simultaneously so that an input is processed minimally on average to predict its label correctly.  This leads us to combine both losses in a single loss function: 

\begin{equation}
\mathcal{L}=\mathcal{L}_\mathrm{MC}+\lambda \mathcal{L}_\mathrm{Cost}
\label{eq:general_loss}
\end{equation} 

\noindent where $\mathcal{L}_\mathrm{MC}$ is the multi-class classification loss, $\mathcal{L}_\mathrm{Cost}$ is the computational cost and $\lambda$ is a trade-off parameter between accuracy and cost. 

Let $\mathbf{\hat{y}}_{i}$ be the classification vector output by the $i^\mathrm{th}$ early-exit block and $c_{i}$ be the computational cost of the network, measured in number of floating-point operations (FLOPs), up to this early-exit block. The inference procedure (Section \ref{subsection:inference}) dictates the following final classification output vector: 

\begin{equation}
\begin{aligned}
\mathbf{\hat{y}} = \; & \mathrm{I}_{ \{ h_{0} \geq T \} }  \mathbf{\hat{y}}_{0} + \mathrm{I}_{ \{ h_{0}<T \} } \{ \\ 
& \mathrm{I}_{\{h_{1} \geq T \}} \mathbf{\hat{y}_{1}} + \mathrm{I}_{\{h_{1}<T \}} \{ \dots \\ 
& \mathrm{I}_{\{h_{N-1} \geq T \}} \mathbf{\hat{y}}_{N-1} + \mathrm{I}_{\{h_{N-1}<T \}} \mathbf{\hat{y}}_{N} \}\dots\} \\
\end{aligned}
\label{eq:non_differential}
\end{equation}

\noindent where $\mathrm{I}_{\{\cdot\}}$ is the indicator function and $N$ is the number of early-exit blocks. $\hat{y}_N$ denotes the final softmax  output of the CNN (it is not the output of an early-exit block). 

We cannot directly use the expression in Eq. \eqref{eq:non_differential} for training because it is not differentiable due to the indicator functions. The indicator function can be approximated by the sigmoid function, and because our confidence scores ($h_i$) are  produced by sigmoid activation functions, we obtain the following \emph{soft} classification output vector: 

\begin{equation}
\begin{aligned}
\mathbf{\hat{Y}}_0 = \; & h_{0} \mathbf{\hat{y}}_{0} + (1 - h_{0}) \{ \\
& h_{1} \mathbf{\hat{y}}_{1} + (1 - h_{1}) \{ \dots \\ 
& h_{N-1} \mathbf{\hat{y}}_{N-1} + (1 - h_{N-1}) \mathbf{\hat{y}}_{N} \}\dots\} \\
\end{aligned}
\label{eq:req_y}
\end{equation}

\noindent which can be more conveniently expressed as a recursive formula: 
\begin{equation}
\begin{aligned}
\mathbf{\hat{Y}}_{i} = \; & h_{i} \mathbf{\hat{y}}_{i} + (1 - h_{i}) \mathbf{\hat{Y}}_{i+1} \;\;\;\; \forall i=0,1,\dots, N-1 \\
\mathbf{\hat{Y}}_{N} = \; & \mathbf{\hat{y}}_{N}.
\end{aligned}
\end{equation}

We can similarly define the \emph{soft} version of the computational cost as: 

\begin{equation}
\begin{aligned}
C_{i} = \; & h_{i} c_{i} + (1 - h_{i}) C_{i+1} \\
C_{N} = \; & c_{N} \\ 
\end{aligned}
\label{eq:req_c}
\end{equation}

\noindent where $c_N$ denotes the computational cost of the whole network from start to the final softmax output. 

Given the definitions above, we can finally write the first version of our loss function: 

\begin{equation}
\begin{aligned}
\mathcal{L}_\mathrm{v1} &= \mathrm{CE}(\mathbf{y},\mathbf{\hat{Y}}_{0}) + \lambda C_{0}
\end{aligned}
\label{eq:loss_v1}
\end{equation}

\noindent where $\mathrm{CE}(\cdot)$ is the cross-entropy loss and $\mathbf{y}$ denotes the ground-truth label. 

The problem with $\mathcal{L}_\mathrm{v1}$ is that, due to the recursive natures of $\mathbf{\hat{Y}}_{0}$ and $C_0$, the later an early-exit block, the smaller its contribution to the overall loss. To see this, consider the multiplicative factor of $\mathbf{\hat{y}}_N$ in Eq. \eqref{eq:req_y}: $\prod_{i=0}^{N-1} (1-h_i)$. Since each $h_i \in [0,1]$, as $i$ grows (i.e. going deeper), the contribution of early-exit block $i$ to the overall loss goes down, consequently, it receives less and less supervisory signal. In our experiments, we observed that EENets trained using $\mathcal{L}_\mathrm{v1}$ showed little diversity in the exit blocks preferred by the inputs and an early stage exit-block  (small $i$) was dominant. Hence, EENets trained with $\mathcal{L}_\mathrm{v1}$ performed poorly. 

To address the shortcoming of $\mathcal{L}_\mathrm{v1}$, we consider the exit block from which the input would possibly not exit as a latent variable and minimize an expected loss over it. 

Suppose for  a specific input, we knew upto which early-exit block it would {\bf not} exit. For example, if we knew that a specific input would exit at the final output of the CNN (therefore, it will not exit from any of the early-exit blocks), then, for this example, it would be sufficient to consider  only the loss term related to $\mathbf{\hat{y}}_N$, and ignore the loss term related to earlier exits. Similarly, if we knew that an example would not exit from early-exit block 0, then we would not add the losses related to this block into the overall loss. 

However, we do not apriori know from which early-exit block a specific example would exit (or not exit). For this reason, we can consider the index of the block from which the example would pass (without exiting) as a latent variable. If we assume a uniform prior over all exit blocks, minimizing the expected value of the loss over this latent variable, we arrive at: 

\begin{equation}
\begin{aligned}
\mathcal{L}_\mathrm{v2} &= \sum_{i=0}^{N} (\mathrm{CE}(\mathbf{y},\mathbf{\hat{Y}}_{i}) + \lambda C_{i}).
\end{aligned}
\label{eq:loss_v2}
\end{equation}

We propose to use $\mathcal{L}_\mathrm{v2}$ to train EENets, where each early-exit block has a chance to contribute to the loss and, hence, receive supervision signal.  
%prediction $\mathbf{\hat{Y}}_{i}$ (relatively the cross-entropy loss) since the equation of the prediction starts with its confidence score $h_{i}$ that cannot be dominated by previous confidence scores. Even if some of the exit blocks are dominant with high confidence scores, the others can promote the loss function and can be properly back-propagated as well.

Overall, all exit blocks contribute to the loss function for all examples, even if easy examples can be classified at the earlier early-exit blocks. Multiple outputs coming from all exit blocks are trained jointly in $\mathcal{L}_\mathrm{v2}$.  Thanks to our novel loss function, $\mathcal{L}_\mathrm{v2}$, (i) EENets avoid the dead layer problem occurring in many previous work \citep{msdnet, branchynet, sacrificing}, and (ii) EENets do not require a complicated multi-stage training process neither.

%Finally, note that the cost loss $\mathcal{L}_{Cost}$ is an important factor to avoid entire execution of network for all examples. As seen from Eq. \eqref{eq:req_y}, because the most possible correct predictions come from the last exit block that exploits all capacity of the model, it is expected that the confidence scores tend to be higher value at the deeper exit points, gradually. However, the cost penalty $\mathcal{L}_{Cost}$ forces the model to terminate early since the shallower early-exit blocks have less computational cost. Therefore, the confidence scores of the shallower early-exit blocks are promoted. In the big picture, the confidence scores actually learn if the predictions of the following exit blocks are worth executing more layers with the additional cost coming from these layers.

\begin{figure}
\centering
\includegraphics[width=0.48\textwidth]{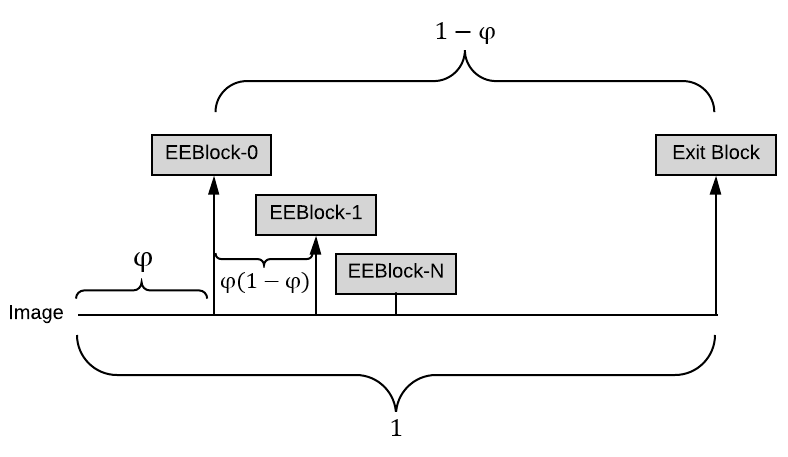} \\
\includegraphics[width=0.48\textwidth]{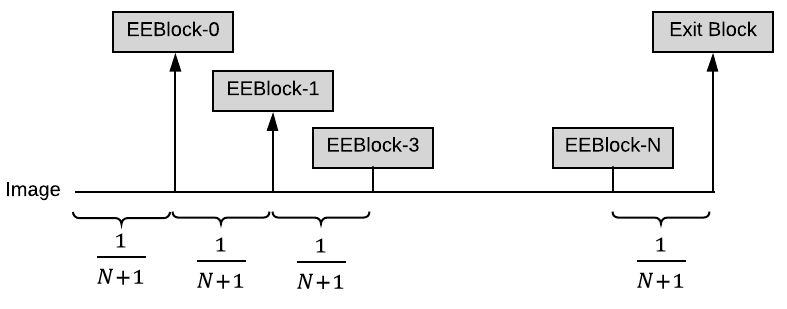} \\
\caption[Distributing early-exit blocks to a network]{Distributing early-exit blocks to a network. \textit{Pareto}, \textit{Golden Ratio} and \textit{Fine} can be represented in the \textbf{upper} figure. The $\varphi$ denotes the ratio used in the methods. For example, $\varphi$ will be $0.2$, $0.6180$ and $0.05$ for \textit{Pareto}, \textit{Golden Ratio} and \textit{Fine} distributions, respectively. $N$ shows the number of early-exit blocks. The \textbf{below} figure shows the \textit{Linear} distribution where the computational costs between consecutive early-exit blocks are same and this cost can be calculated by the desired number of the early-exit blocks. Notice that the total cost is represented by 1 since our cost terms are rates (i.e. $c_i \in [0, 1]$).}
\label{fig:distribution}
\end{figure}

\subsection{Distributing Early-exit Blocks to a Network}
\label{subsection:distributing-blocks}
The number of early-exit blocks (EE-blocks) and how they are distributed over the base CNN are other important factors in the architecture of EENets. The additional parameters introduced by the EE-blocks are very small and usually negligible (e.g. $\sim$0.0002\% of the total parameters of EENet-110 for each EE-block on a 10-class datasets), the early-exit blocks can be added as much as desired. %The number of EE-blocks depends on the depth of the network. 

We propose five different ways for determining where a given number of EE-blocks should be added in a given base CNN: (i) Pareto, (ii) Golden Ratio, (iii) Fine, (iv) Linear and (v) Quadratic. According to the Pareto principle, 80\% of the results come from 20\% of the work. Our  \textit{Pareto} distribution is inspired by this principle: the first EE-block splits the network according to the Pareto principle where 20\% of the total computational cost is calculated in terms of the number of floating-point operations (FLOPs). Similarly, the second EE-block splits the rest of the network (i.e. starting right after the first EE-block until the end) again into 20\%-80\%. This pattern continues until all EE-blocks are added. In the \textit{Fine} distribution method, each EE-block divides the network at 5\%-95\% based on the total FLOPs. The \textit{Golden ratio} distribution uses the golden ratio, 61.8\%-38.2\%.

The \textit{Linear} and \textit{Quadratic} distributions split the network in such a way that the computational cost of the layers between two consecutive EE-blocks increases in linear or quadratic form, respectively. Figure \ref{fig:distribution} illustrates some of the distribution methods. Note that there is not a best distribution method for all EENets or datasets. The effects of the distribution method used should be observed empirically on the specific problem.

\section{Experiments}
\label{section:b4}

In our experiments, we chose ResNets \citep{resnet} as our base CNNs for their widespread use (although, our early-exit blocks can be applied to any CNN architecture). We obtained  early-exit (EE) versions of ResNets and compared their performance to that of non-EE (i.e. original) versions on MNIST \citep{mnist}, CIFAR10 \citep{cifar}, SVHN \citep{svhn} and Tiny-ImageNet \citep{tiny-imagenet} datasets. In addition to ResNets, we also experimented with a small, custom CNN on the MNIST dataset. 

The experiments are diversified in order to observe the effects of EENets in different aspects and certain conditions. In this section, we try to answer the following questions through comprehensive experiments:

\begin{itemize}
\item Do EENets really work? That is, is the inference process terminated for individual examples at different exit locations? Is there a variety in the exit locations chosen per example by the network?
\item Are EENets successful when compared to their counterparts in terms of  computational cost and  accuracy?
\item Which type of early-exit block yield better results?
\item How does the distribution of early-exit blocks affect the accuracy and the computational cost?
\end{itemize}

We conducted our experiments on a machine with a i7-6700HQ CPU processor with 16GB RAM and two NVIDIA Tesla PICe P100 16GB. We implemented EENets both in two different frameworks (Keras (v2.1.5) and PyTorch (v1.0.1)) to verify behavior and performance. We chose PyTorch for its flexibility\footnote{For example, it is not straightforward in Keras to write different feed-forward functions for training and testing. During training, each example passes through the whole network, whereas during testing, execution might stop in any of the early-exit blocks due to hard-thresholding.}. 
%The models are implemented in Keras with the Tensorflow backend but we also implemented a toy model in PyTorch. 
Both  Keras and PyTorch implementations are available at GitHub\footnote{PyTorch: \url{https://github.com/eksuas/eenets.pytorch}, Keras: \url{https://github.com/eksuas/eenets.keras}}. Unless otherwise noted, all results reported in this section were produced by the PyTorch code.  

In MNIST \citep{mnist} and Tiny-ImageNet \citep{tiny-imagenet} experiments, the models were optimized by Adam with  \textit{learning rate $ = 0.001$}. The mini-batch size in the experiments was $32$. Most of the models were trained up to $200$ epochs unless otherwise stated.

\begin{sloppypar}
On SVHN \citep{svhn} and CIFAR10 datasets \citep{cifar}, we trained the models using the configurations  given in the ResNet paper \citep{resnet}. In these experiments, we used SGD with a mini-batch size of $256$. The learning rate starts from $0.1$ and is divided by $10$ per $100$ epochs. The models were trained for up to $350$ epochs. We used  \textit{weight decay $ = 0.0001$} and  \textit{momentum $ = 0.9$}.
\end{sloppypar}

\begin{figure}
\centering
\parbox{7cm}{\includegraphics[width=7cm]{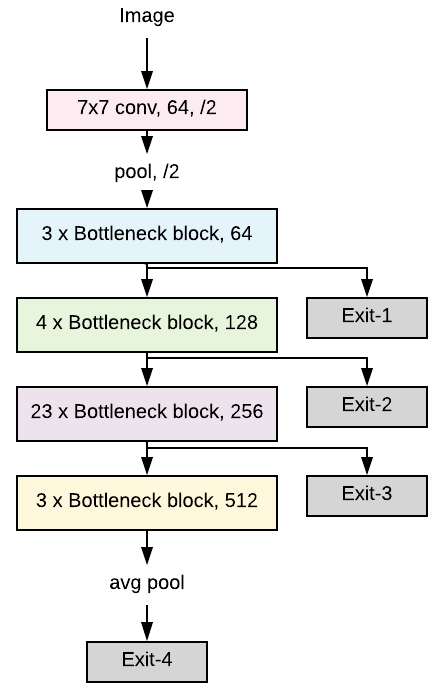}}
\caption[EENet-101 consisting of three \textit{Pool}-type EE-blocks]{EENet-101 model with three early-exit blocks. The \textit{Pool}-type of early-exit blocks and the bottleneck blocks are employed. The architecture is in the form of Naive ResNet models.}
\label{fig:eenet101}
\end{figure}

\begin{figure}
\centering
\parbox{7cm}{\includegraphics[width=7cm]{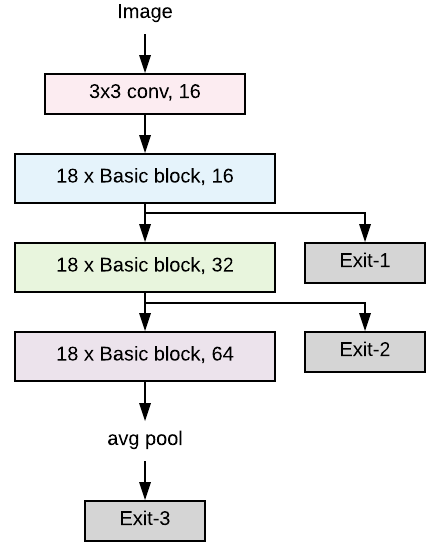}}
\caption[EENet-110 consisting of two \textit{pool}-type EE-blocks]{EENet-110 model with two early-exit blocks. The early-exit blocks are in the \textit{pool}-type. It is another example of the architectures in the 6n+2 ResNet form.}
\label{fig:eenet110}
\end{figure}

\subsection{Experimented Architectures}
We added early-exit blocks to ResNet models \citep{resnet} with both basic and bottleneck architectures. The early-exit (EE) ResNets based on bottleneck architectures consist of 50, 101 and 152 layers. By modifying the 6n+2 layers ResNet models \citep{resnet}, we have constructed 20, 32, 44 and 110 layers EENets. Various numbers of early-exit blocks were distributed based on  the capacity of the models. The models which have a large capacity were trained on CIFAR10 \citep{cifar} and SVHN \citep{svhn}. In addition, the early-exit version of Naive\footnote{The ResNet paper \citep{resnet} introduced two types of models: \textit{Naive} and \textit{6n+2 layer} types. The larger capacity \textit{Naive} models were designed for the ImageNet dataset and they consist of four main stages. The other, \textit{6n+2 layers} type, has three main stages whose total number of layers equal to 6n+2 for integer n.} ResNets such as EENet-18 were evaluated on Tiny-ImageNet \citep{tiny-imagenet}.

On the other hand, the models having a smaller capacity were evaluated on MNIST \citep{mnist} to observe how EENets perform in the situation of a dataset forcing the capacity of the model. These small capacity networks are composed of 6, 8 and 18 layers with a small number of filters. 

Some of the ResNet based architectures that are evaluated in our experiments are shown in Figures \ref{fig:eenet101} and \ref{fig:eenet110}. Our own design EENet-8  is a very small CNN having between 2 to 8 filters in its layers. We ran this low capacity model on the MNIST dataset.

%\subsection{Metrics}
%The testing accuracy of EENets is measured by considering the predicted labels at exit locations. To measure computational cost, we use the number of floating-point operations (FLOPs).
%label at the point where the example is classified and the execution is terminated. For each example, our custom accuracy metric keeps the prediction and the computational cost consumption of the first exit block that satisfies the confidence threshold for that example. Later on, the total testing accuracy of the model and the computational cost are calculated by using the accuracy of that saved class labels and their costs.

\begin{table*}
\centering
\begin{tabular}{|l||c|c|c|}
\multicolumn{4}{c}{A. Exit distribution on EENet-8-$\mathcal{L}_{v2}$}\\
\hline
\multirow{2}{*}{\bf Exit Blocks} & 
\multirow{2}{*}{\bf FLOPs} & \bf Relative & \bf \#examples \\
& & \bf Cost & \bf that exit \\
\hline
\hline
EE-block0 &  546 & 0.08 & 47  \\
EE-block1 & 1844 & 0.26 & 2247 \\
Last Exit  & 6982 & 1.00 & 7706 \\ 
\hline
\end{tabular}
\\
\begin{tabular}{|l||r|r|r|c|c|}
\multicolumn{6}{c}{}\\
\multicolumn{6}{c}{B. Benchmark of different loss functions}\\
\hline
\multirow{2}{*}{\bf Model} & \multicolumn{3}{c|}{\textbf{\# examples that exit from}} & \multirow{2}{*}{\bf Accuracy} & \bf Relative \\
\cline{2-4}
& \bf EE-block0 & \bf EE-block1 & \bf Last Exit & & \bf Cost \\
\hline
\hline
ResNet-8                     & -     & -    & 10000 & 97.38 & 1.00 \\
EENet-8-$\mathcal{L}_{MC}$   & 0     & 0    & 10000 & 97.42 & 1.00 \\
EENet-8-$\mathcal{L}_{Cost}$ & 10000 & 0    & 0     & 10.32 & 0.08 \\
EENet-8-$\mathcal{L}_{v1}$   & 6614  & 3386 & 0     & 54.05 & 0.14 \\
EENet-8-$\mathcal{L}_{v2}$   & 47    & 2247 & 7706  & 96.55 & 0.82 \\
\hline
\multicolumn{6}{c}{}\\
\end{tabular}
\caption[Exit distribution of the MNIST examples with different loss functions]{\textbf{Exit distribution of the MNIST examples with different loss functions.} This table shows the results of MNIST examples evaluated on the EENet-8 model with 20 epochs. In the \textbf{upper} table, the computational cost rates and the number of FLOPs from the beginning to the early-exit blocks are given in the \textbf{\textit{Relative Cost}} and \textbf{\textit{FLOP}} columns, respectively. \textbf{\textit{\# examples that exit}} shows the number of examples that are classified at that exit block. The exit distribution of the MNIST test examples (10000 examples) on the EENet-8 models are shown in the \textbf{lower} table where EENet-8-$\mathcal{L}_{v1}$, EENet-8-$\mathcal{L}_{v2}$,  EENet-8-$\mathcal{L}_{Cost}$ and EENet-8-$\mathcal{L}_{MC}$ are the EENet-8 models trained with only the $\mathcal{L}_{v1}$, $\mathcal{L}_{v2}$, $\mathcal{L}_{Cost}$ and $\mathcal{L}_{MC}$ loss functions, respectively. \textbf{\textit{Last Exit}} represents the last exit block. Testing accuracy is given in the \textbf{\textit{Accuracy}} column. Note that the cost of ResNet-8 is always 1 since it computes the whole model. Since early-exit blocks are not available for ResNets, “-” is placed in the early-exit columns.}
\label{table:exitDist}
\end{table*}

\subsection{Results on MNIST}
First, we performed a set of experiments on the MNIST dataset \citep{mnist} to see if the confidence scores  are meaningful (i.e. they are related to  the accuracy of the predictions made by  these early-exit blocks) and if they  have a variety in inputs. In these basic tests, the EENet-8 model was employed with quadratically distributed two \textit{Pool}-type early-exit blocks whose number of floating-point operations (FLOPs) and costs are given in Table \ref{table:exitDist}. The exit distribution of the MNIST examples on the EENet-8 models trained with different loss functions are given in Table \ref{table:exitDist} as well. We used  $\lambda=1$  in these experiments.

As expected, the model EENet-8-$\mathcal{L}_\mathrm{Cost}$ terminates the executions at the first early-exit block by considering only the computational cost while EENet-8-$\mathcal{L}_\mathrm{MC}$ classifies all examples at the last exit block to get the highest accuracy. On the other hand, the EENet-8 model trained with $\mathcal{L}_\mathrm{v2}$ takes both the cost and accuracy into account; as a consequence, it maintains the accuracy by spending less computational cost ($0.82$ of the original). EENet-8-$\mathcal{L}_\mathrm{v1}$ performs poorly as expected (see Section \ref{subsection:training} for the discussion). Moreover, we observe that examples exit at a variety of locations (Table \ref{table:exitDist}): 47, 2247 and 7706 numbers of test examples of MNIST are classified at the early-exit (EE) block-0, EE-block-1 and the last exit layer of the EENet-8 model, respectively. This experiment shows that our loss function,$\mathcal{L}_\mathrm{v2}$, performs as expected and maintains the balance between the accuracy and computational cost. 
%Because of that, the EENet models trained with the $\mathcal{L}_{v2}$ loss function are evaluated in the remaining experiments. 

\begin{table*}
\centering
\begin{tabular}{|c||c|c|c|r|r|r|}
\hline
\multirow{2}{*}{\boldmath$\lambda$} &
\multirow{2}{*}{\bf Accuracy} &
\bf Time &
\bf Relative & 
\multicolumn{3}{c|}{\textbf{\# examples that exit from}} \\
\cline{5-7}
& & \bf ($\mu$s) & \bf Cost & \bf EEB-0 & \bf EEB-1 & \bf Last Exit \\
\hline
\hline
0.50 & 98.84 & 638.0 & 1.00 & 0    & 0    & 10000 \\
0.70 & 98.52 & 711.6 & 1.00 & 0    & 0    & 10000 \\
0.90 & 97.46 & 681.4 & 0.78 & 1120 & 1498 & 7382  \\
0.95 & 97.48 & 643.3 & 0.74 & 1230 & 1883 & 6887  \\
1.00 & 96.22 & 615.0 & 0.82 & 359  & 1879 & 7762  \\
1.05 & 97.53 & 593.6 & 0.85 & 168  & 1771 & 8061  \\
1.10 & 98.34 & 641.9 & 0.85 & 3    & 1939 & 8058  \\
1.15 & 85.93 & 557.8 & 0.45 & 48   & 7245 & 2707  \\
1.30 & 86.96 & 487.3 & 0.26 & 0    & 9996 & 4     \\
1.50 & 85.19 & 476.4 & 0.26 & 0    & 9997 & 3     \\
\hline
\multicolumn{7}{c}{} 
\end{tabular}
\caption[Effects of $\lambda$ trade-off on the loss functions $\mathcal{L}_{v2}$]{\textbf{Effects of \boldmath$\lambda$ trade-off on the loss functions \boldmath$\mathcal{L}_{v2}$.} EENet-8 model starting with 4 filters is evaluated on MNIST. The model is trained with different $\lambda$ trade-off by using ADAM optimizer on 20 epochs. The exit distribution of the MNIST test examples (10000 examples) is shown in the table where the values are the results of the last epoch. Testing accuracy is given in the \textbf{\textit{Accuracy}} column. \textbf{\textit{Time}} column shows the average wall clock time of inference procedure in microseconds ($\mu$s). The computational cost rates are given in the \textbf{\textit{Relative Cost}} column. \textbf{\textit{$\#$ examples that exit from}} shows the number of examples that are classified at that exit block.}
\label{table:lambda_mnist}
\end{table*}

%Note that the success of our model (in Table \ref{table:exitDist}) in terms of the computational cost is not as good as the results of experiments performed on other models and datasets (the experiments will be given in the following sections). The reason behind that, the capacity of the model EENet-8 is very low than the traditional ResNets in the original paper \citep{resnet}. As seen from Table \ref{table:exitDist}, the FLOPs of EENet-8 is 6982 while a traditional ResNet110 has 256.32 (MMac) FLOPs. We deliberately choose this model to force it to early classify the examples of MNIST as an easy dataset. Otherwise, with a large capacity model, MNIST examples are classified in the early stages of the model (mostly at the first early-exit block) because the dataset consists of simple examples. As a consequence, the proportional distribution may not be examined on MNIST if the model capacity is large like a traditional ResNet. 

\begin{figure*}
\begin{minipage}{.46\textwidth}
\centering
\includegraphics[width=\textwidth]{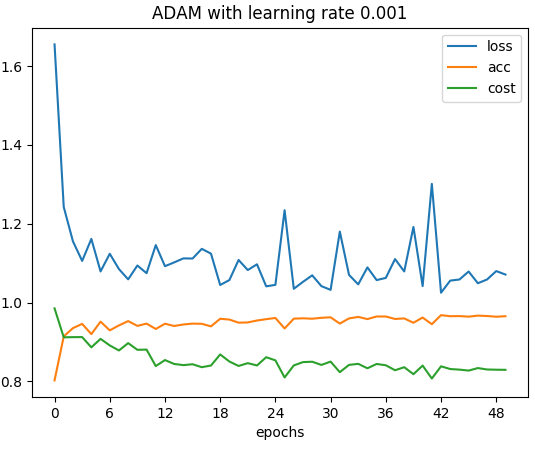} \\
\includegraphics[width=\textwidth]{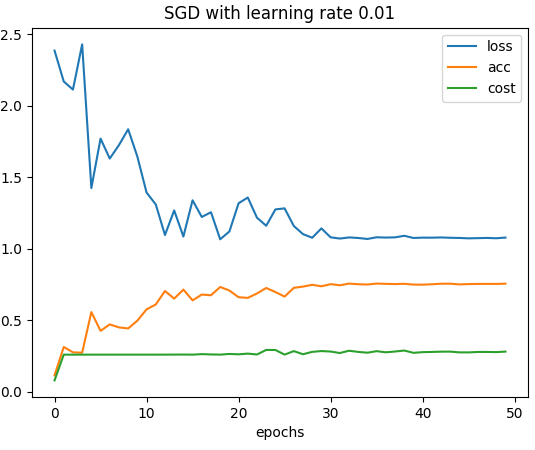} \\
\end{minipage}
\qquad
\begin{minipage}{.46\textwidth}
\centering
\includegraphics[width=\textwidth]{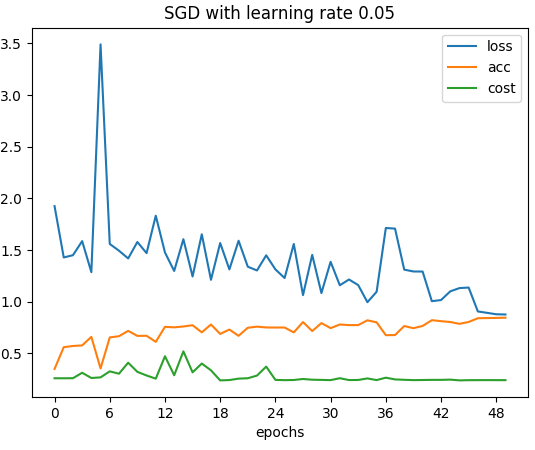} \\
\includegraphics[width=\textwidth]{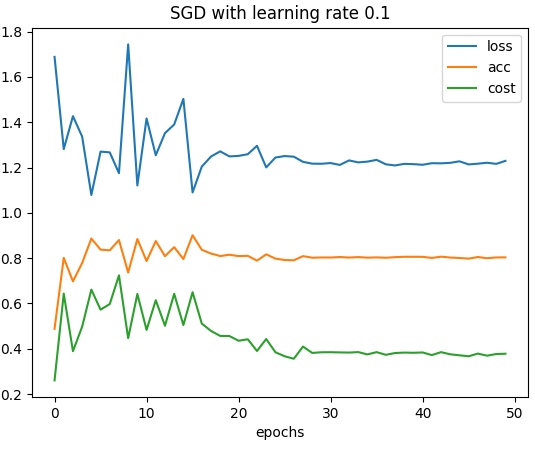} \\
\end{minipage}
\caption{Epochs vs accuracy (orange), loss (blue) and computational cost (shown with green color) of the EENet-8 model on MNIST are given above where the values of accuracy and cost are $\in [0,1]$.}
\label{fig:optimizer}
\end{figure*}

The computational cost, accuracy and loss values per epoch are shown in Figure \ref{fig:optimizer}. We evaluate the model with different optimizers and learning rates. Adam optimizer with learning rate 0.001 gives the best results. %As seen in the figure, the accuracy increases while the computational cost decreases gradually. As a consequence, we continue training the models with Adam in the remaining experiments.

\begin{figure}
\centering
\includegraphics[width=.48\textwidth]{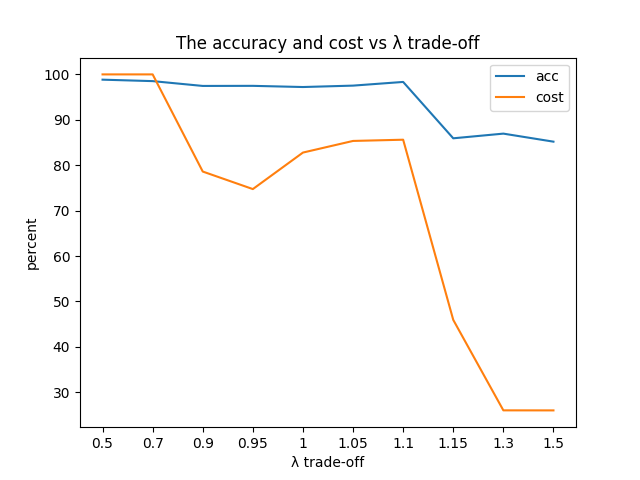}
\caption{The accuracy and computational cost vs $\lambda$ trade-off on EENet-8.}
\label{fig:lambda}
\end{figure}

We performed another set of experiments  on MNIST to observe the effects of $\lambda$ trade-off on the loss function $\mathcal{L}_{v2}$. The results are presented in Table \ref{table:lambda_mnist} and Figure \ref{fig:lambda}. The best balance between the accuracy and the computational cost is observed with $\lambda 0.95$. However, the effects of $\mathcal{L}_\mathrm{MC}$ or $\mathcal{L}_\mathrm{Cost}$ can be changed through the $\lambda$ trade-off if more accurate results or less computational cost consumption are desired (e.g. $\lambda$ can be decreased to obtain more accurate results if the computational cost is not an issue).

\begin{figure*}
\centering
\includegraphics[width=.8\textwidth]{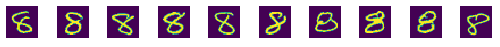}
\caption{Randomly sampled MNIST examples that exit from the first early-exit block.}
\label{fig:exit0}
\centering
\includegraphics[width=.8\textwidth]{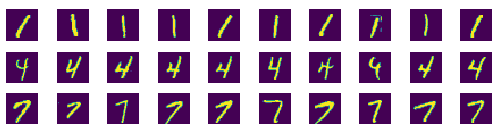}
\caption{Randomly sampled MNIST examples that exit from  the second early-exit block.}
\label{fig:exit1}
\centering
\includegraphics[width=.8\textwidth]{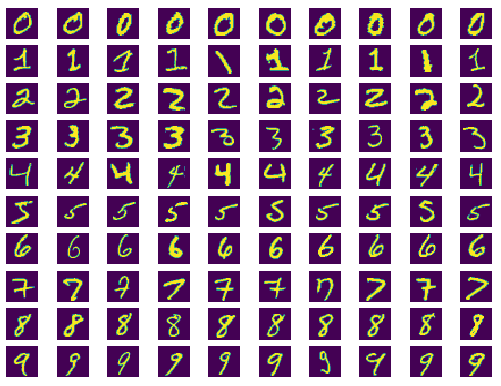}
\caption{Randomly sampled MNIST examples that exit from  the base CNN.}
\label{fig:exit2}
\end{figure*}

Random MNIST examples classified with EENet-8 which consists of two early-exit (EE) blocks are shown in Figures \ref{fig:exit0}, \ref{fig:exit1} and \ref{fig:exit2} as classified at the EE-block-0, EE-block-1 and the last exit blocks of the model, respectively. We observe that the early-exit blocks are specialized in visually similar examples of the same class or in a few visually similar classes. For example, the EE-block-0 is only specialized in the class of the digit eight in this model (i.e. visually similar examples of the same class). On the other hand, the EE-block-1 seems to be specialized in digits   one, four and seven. Note that these classes are visually similar as well.

Finally, we tested different distributions of the early-exit blocks. %It is not hard to classify MNIST dataset using ResNet based models since these models enjoy a large capacity than MNIST required. 
We observe that keeping the first early-exit blocks in the very beginning of the model decreases the cost excessively. Due to this, the \textit{Quadratic} distribution with a small number of EE-blocks can be a good choice in this situation.% (i.e. a large-capacity model with a simple dataset that can be easily classified).

\begin{table}
\centering
\begin{tabular}{|l||c|c|}
\hline
\multirow{2}{*}{\bf Block Type} &
\multirow{2}{*}{\bf Accuracy} &
\bf Relative \\
& & \bf Cost \\
\hline
\hline
Plain  &  90.05 & 0.01 \\
Pool   &  94.59 & 0.06 \\
Bnpool &  94.45 & 0.06 \\
\hline
\multicolumn{3}{c}{}
\end{tabular}
\caption[Types of the early-exit blocks]{\textbf{Types of the early-exit blocks.} The type of early-exit blocks: \textit{Plain}, \textit{Pool} and \textit{Bnpool} are evaluated within the  EENet-50 model on the SVHN dataset. \textit{(Results in this table were obtained using the Keras code.)}
%\textit{Pool}-type is the more accurate one with a small difference.
}
\label{table:exitType}
\end{table}

\subsection{Results on Tiny-ImageNet}

We evaluated EENet-18 on the Tiny-ImageNet dataset \citep{tiny-imagenet} which consists of  200 classes with 500 training and 50 validation images (down-sampled to 64-by-64) per class,  from the original ImageNet dataset \citep{imagenet}. EENet-18 accuracies together with average wall-clock time during inference for varying $\lambda$ are given in Table \ref{table:lambda_tiny}. As expected, the average wall clock time is inversely correlated with  the relative  computational cost rates. On interesting  observation is that on Tiny-ImageNet, higher values  of  $\lambda$  yield better accuracies, although, higher  $\lambda$ values give more importance to the $\mathcal{L}_\mathrm{Cost}$ component of the overall loss function ($\mathcal{L}_\mathrm{v2}$).  

\begin{table}
\centering
\begin{tabular}{|c||c|c|}
\hline
\multirow{2}{*}{\bf Model} & 
\multirow{2}{*}{\bf Accuracy} & \bf Time \\
& & \bf ($\mu$s) \\
\hline
\hline
ResNet-18 & 38.98 & 557.6 \\
\hline
\end{tabular}

\begin{tabular}{|c||c|c|c|}
\multicolumn{4}{c}{}\\
\hline
\multirow{2}{*}{\boldmath$\lambda$} &
\multirow{2}{*}{\bf Accuracy} & \bf Time & \bf Relative \\
& & \bf ($\mu$s) & \bf Cost \\
\hline
\hline
0.4 & 38.96 & 521.7 & 0.99 \\
0.6 & 38.63 & 553.7 & 0.98 \\
0.7 & 39.08 & 503.3 & 0.91 \\
0.8 & 39.47 & 411.0 & 0.70 \\
0.9 & 39.61 & 389.6 & 0.67 \\
1.0 & 40.78 & 338.8 & 0.51 \\
1.1 & 40.39 & 336.1 & 0.48 \\
1.2 & 41.21 & 316.3 & 0.43 \\
1.3 & 41.40 & 286.3 & 0.44 \\
1.5 & 41.49 & 317.2 & 0.42 \\
1.7 & 41.75 & 322.7 & 0.42 \\
1.9 & 41.15 & 328.4 & 0.42 \\
\hline
\end{tabular}

\begin{tabular}{|c||c|c|c|}
\multicolumn{4}{c}{}\\
\hline
\multirow{2}{*}{\boldmath$\lambda$} &
\multirow{2}{*}{\bf Accuracy} & \bf Time & \bf Relative \\
& & \bf ($\mu$s) & \bf Cost \\
\hline
\hline
0.4 & 42.63 & 515.4 & 67.44 \\
0.5 & 42.49 & 488.1 & 65.64 \\
0.6 & 42.28 & 496.2 & 65.47 \\
0.7 & 42.84 & 464.1 & 64.69 \\
0.8 & 43.28 & 505.4 & 64.69 \\
0.9 & 42.84 & 473.0 & 64.66 \\
1.0 & 43.00 & 487.9 & 64.67 \\
1.1 & 43.50 & 474.9 & 64.46 \\
1.2 & 43.45 & 478.6 & 64.32 \\
1.3 & 42.85 & 469.1 & 64.25 \\
1.4 & 43.46 & 462.7 & 64.28 \\
1.5 & 43.36 & 454.1 & 62.47 \\
\hline
\multicolumn{4}{c}{}
\end{tabular}
\caption[Effects of $\lambda$ trade-off on the loss functions $\mathcal{L}_{v2}$]{\textbf{Effects of \boldmath$\lambda$ trade-off on the loss functions \boldmath$\mathcal{L}_{v2}$.} Tiny-ImageNet dataset is evaluated on ResNet-18 (\textit{upper table}). It is also evaluated on EENet-18 model with 3 (\textit{middle table}) and 5 (\textit{lower table}) EE-blocks. The models are trained with different $\lambda$ trade-off by using ADAM optimizer on 20 epochs. Validation accuracy, average wall clock time and relative computational cost rates (according to ResNets) are given above.}
\label{table:lambda_tiny}
\end{table}

\subsection{Results on SVHN}
%We evaluated our proposed model with the original ResNets on SVHN dataset \citep{svhn} as well.

On the SVHN dataset \citep{svhn}, we  first examined the performance of different types of  early-exit blocks on the EENet-50 model. The results are given in Table \ref{table:exitType}. The \textit{Pool}-type early-exit block produces the most accurate results with a very small margin over the  \textit{Bnpool}-type. Consequently, we employed the \textit{Pool}-type early-exit blocks in the rest of the experiments.

\begin{table}
\centering
\begin{tabular}{|l||c|c|c|}
\hline
\multirow{2}{*}{\bf Model} &
\multirow{2}{*}{\bf Accuracy} & 
\multirow{2}{*}{\bf \#Params} &
\bf FLOPs \\
& & & \bf (MMac) \\
\hline
\hline
ResNet-20  & 95.61 & 0.27M & 41.41 \\
ResNet-32  & 95.72 & 0.47M & 70.06 \\
ResNet-44  & 95.79 & 0.67M & 98.72 \\
ResNet-110 & 95.68 & 1.74M & 256.32 \\
\hline
\multicolumn{4}{c}{}
\end{tabular}
\caption[Results of 6n+2 based ResNets on SVHN]{\textbf{Results of 6n+2 based ResNets on SVHN.} The average number of floating-point operations are given in the column of \textbf{\textit{FLOP}}. \textbf{\textit{\#Params}} denotes the total number of model parameters. The given accuracy in the table is the testing accuracy.}
\label{table:svhn6resnet}
\end{table}

\begin{table*}
\centering
\begin{tabular}{l||l||c|c|l|c|}
\cline{2-6}
&
\multirow{2}{*}{\bf Model} & 
\multirow{2}{*}{\bf Accuracy} & 
\bf \# Early-exit & 
\multirow{2}{*}{\bf Cost Percent of EE-Blocks}
& \bf Relative \\
& & & \bf Blocks & & \bf Cost\\
\hline
\hline
\multicolumn{1}{|l||}{\multirow{4}{*}{\begin{turn}{90}Fine\end{turn}}}
& EENet-20  & 93.74 & 3  & 13,24,36                     & 1.00 \\
\multicolumn{1}{|l||}{}
& EENet-32  & 94.30 & 5  & 8,14,21,28,35                & 0.87 \\
\multicolumn{1}{|l||}{}
& EENet-44  & 94.43 & 6  & 5,10,15,20,25,30             & 0.77 \\
\multicolumn{1}{|l||}{}
& EENet-110 & 94.46 & 10 & 6,11,15,19,23,27,30,34,37,41 & 0.30 \\
\hline
\hline
\multicolumn{1}{|l||}{\multirow{4}{*}{\begin{turn}{90}Pareto\end{turn}}}
& EENet-20  & 93.65 & 3  & 24,36,57                      & 0.93 \\
\multicolumn{1}{|l||}{}
& EENet-32  & 94.22 & 5  & 21,40,54,61,68                & 0.60 \\
\multicolumn{1}{|l||}{}
& EENet-44  & 95.06 & 6  & 20,38,53,63,67,76             & 0.62 \\
\multicolumn{1}{|l||}{}
& EENet-110 & 95.54 & 10 & 21,37,50,60,69,74,80,83,87,91 & 0.83 \\
\hline
\hline
\multicolumn{1}{|l||}{\multirow{4}{*}{\begin{turn}{90}G.Ratio\end{turn}}}
& EENet-20  & 94.12 & 3  & 24,45,68                   & 1.00 \\
\multicolumn{1}{|l||}{}
& EENet-32  & 94.78 & 5  & 14,21,28,40,68             & 0.91 \\
\multicolumn{1}{|l||}{}
& EENet-44  & 94.88 & 6  & 10,15,20,25,38,63          & 0.83 \\
\multicolumn{1}{|l||}{}
& EENet-110 & 95.62 & 10 & 2,4,6,8,10,11,15,25,39,63  & 0.57 \\
\hline
\hline
\multicolumn{1}{|l||}{\multirow{4}{*}{\begin{turn}{90}Linear\end{turn}}}
& EENet-20  & 93.79 & 3  & 36,57,77                      & 0.95 \\
\multicolumn{1}{|l||}{}
& EENet-32  & 94.85 & 5  & 21,35,54,68,86                & 0.86 \\
\multicolumn{1}{|l||}{}
& EENet-44  & 95.02 & 6  & 15,30,43,58,76,90             & 0.90 \\
\multicolumn{1}{|l||}{}
& EENet-110 & 95.55 & 10 & 10,19,28,37,47,56,65,74,83,93 & 0.76 \\
\hline
\multicolumn{6}{c}{}
\end{tabular}
\caption[Results of the 6n+2 based EENets on SVHN]{\textbf{Results of the 6n+2 based EENets on SVHN.} The computational cost rates and the average number of floating-point operations per example are given in the columns of \textbf{\textit{Cost}} and \textbf{\textit{FLOP}}, respectively. The distribution methods of models are given in the first column. \textbf{\textit{\#E}} denotes the number of EE-blocks. \textbf{\textit{Cost Percent of EE-Blocks}} shows the distribution of cost percent of EE-blocks. The given accuracy in the table is the testing accuracy.}
\label{table:svhn6eenet}
\end{table*}

We  evaluated  ResNet 6n+2 architectures \citep{resnet} with four different depths (Table \ref{table:svhn6resnet}) and their  early-exit (EE) counterpart (Table \ref{table:svhn6eenet}) on SVHN. 
EENets achieve %The early-exit (EE) ResNets which have the 6n+2 architecture \citep{resnet} achieve 
similar accuracy as their non-EE versions while reducing the computational cost upto 30\% of the original (\textit{cf.} the cost of the EENet-110 model with Fine distribution in Table \ref{table:svhn6eenet}). 
%\hl{On average, EENets degrade the computational cost to 20\% of that of their non-EE versions.} 
%The results of these experiments are given in Table \ref{table:svhn6resnet} and Table \ref{table:svhn6eenet}. 

We observe that, in general, EENets with ``Fine"ly distributed EE-blocks minimize the computational cost while maintaining accuracy. EENets with \textit{Linear} distribution yield slightly better accuracy than models with Fine distribution, however, they cost more (\textit{cf.} rows corresponding to Linear and Fine in Table \ref{table:svhn6eenet}).
The main reason behind that the first early-exit block of the \textit{Linear} distributed model is located in much deeper layers than the first EE-block of the model of other distributions. For example, the first EE-block of the EENet-32 with \textit{Linear} distribution spends $21\%$ of the total computational cost while the one with \textit{Fine} distribution spends only $8\%$. The other explicit observation is that the computational cost decreases while the model capacity increases in EENet models.

\subsection{Results on CIFAR10}
We tested a variety of EENet models on the CIFAR10 dataset \citep{cifar}. The pattern of results is similar to that of the SVHN dataset. Tables \ref{table:cifar6resnet} and  \ref{table:cifar6eenet} show the results of the 6n+2 and the naive ResNet based models, respectively. 

EE-block versions of 6n+2 architectures achieve similar accuracy with their non-EE counterparts while reducing the computational cost upto 24\% of the original (e.g. the cost of the EENet-110 model with  \textit{Golden Ratio} distribution). As seen in Table \ref{table:cifar6eenet}, the models with  \textit{Golden Ratio} and \textit{Fine} distributions spend less computational costs than the models having other distribution methods. However, their accuracies are not as high as models with the Pareto distribution.

\begin{table}
\centering
\begin{tabular}{|l||c|c|c|}
\hline
\multirow{2}{*}{\bf Model} &
\multirow{2}{*}{\bf Accuracy} & 
\multirow{2}{*}{\bf \#Params} &
\bf FLOPs \\
& & & \bf (MMac) \\
\hline
\hline
ResNet-32  & 93.31 & 0.47M & 70.06 \\
ResNet-44  & 85.46 & 0.67M & 98.72 \\
ResNet-110 & 93.80 & 1.74M & 256.32 \\
\hline
\multicolumn{4}{c}{}
\end{tabular}
\caption[Results of 6n+2 based ResNets on CIFAR10]{\textbf{Results of 6n+2 based ResNets on CIFAR10.} The average number of floating-point operations per example are given in the column of \textbf{\textit{FLOP}}. \textbf{\textit{\#Params}} denotes the total number of model parameters. The accuracy is the testing accuracy.}
\label{table:cifar6resnet}
\end{table}

\begin{table*}
\centering
\begin{tabular}{l||l||c|c|l|c|}
\cline{2-6}
&
\multirow{2}{*}{\bf Model} & 
\multirow{2}{*}{\bf Accuracy} & 
\bf \# Early-exit & 
\multirow{2}{*}{\bf Cost Percent of EE-Blocks}
& \bf Relative \\
& & & \bf Blocks & & \bf Cost\\
\hline
\hline
\multicolumn{1}{|l||}{\multirow{4}{*}{\begin{turn}{90}Fine\end{turn}}}
& EENet-20  & 75.74 & 3  & 13,24,36                     & 0.36 \\
\multicolumn{1}{|l||}{}
& EENet-32  & 78.82 & 5  & 8,14,21,28,35                & 0.28 \\
\multicolumn{1}{|l||}{}
& EENet-44  & 80.94 & 6  & 5,10,15,20,25,30             & 0.28 \\
\multicolumn{1}{|l||}{}
& EENet-110 & 85.93 & 10 & 6,11,15,19,23,27,30,34,37,41 & 0.36 \\
\hline
\hline
\multicolumn{1}{|l||}{\multirow{4}{*}{\begin{turn}{90}Pareto\end{turn}}}
& EENet-20  & 83.59 & 3 & 24,36,57          & 0.56 \\
\multicolumn{1}{|l||}{}
& EENet-32  & 86.34 & 5 & 21,40,54,61,68    & 0.60 \\
\multicolumn{1}{|l||}{}
& EENet-44  & 87.25 & 6 & 20,38,53,63,67,76 & 0.62 \\
\multicolumn{1}{|l||}{}
& EENet-110 & 91.17 & 6 & 21,37,50,60,69,74 & 0.50 \\
\hline
\hline
\multicolumn{1}{|l||}{\multirow{4}{*}{\begin{turn}{90}G.Ratio\end{turn}}}
& EENet-20  & 85.29 & 3 & 24,45,68          & 0.68 \\
\multicolumn{1}{|l||}{}
& EENet-32  & 77.91 & 5 & 14,21,28,40,68    & 0.28 \\
\multicolumn{1}{|l||}{}
& EENet-44  & 78.21 & 6 & 10,15,20,25,38,63 & 0.20 \\
\multicolumn{1}{|l||}{}
& EENet-110 & 84.24 & 6 & 6,10,15,25,39,63  & 0.24 \\
\hline
\hline
\multicolumn{1}{|l||}{\multirow{4}{*}{\begin{turn}{90}Linear\end{turn}}}
& EENet-20  & 83.92 & 3  & 36,57,77                      & 0.56 \\
\multicolumn{1}{|l||}{}
& EENet-32  & 87.00 & 5  & 21,35,54,68,86                & 0.67 \\
\multicolumn{1}{|l||}{}
& EENet-44  & 86.92 & 6  & 15,30,43,58,76,90             & 0.57 \\
\multicolumn{1}{|l||}{}
& EENet-110 & 87.94 & 10 & 10,19,28,37,47,56,65,74,83,93 & 0.38 \\
\hline
\multicolumn{6}{c}{}
\end{tabular}
\caption[Results of the 6n+2 based EENets on CIFAR10.]{\textbf{Results of the 6n+2 based EENets on CIFAR10.} The computational cost rates and the average number of floating-point operations per example are given in the columns of \textbf{\textit{Cost}} and \textbf{\textit{FLOP}}, respectively. The distribution methods of models are given in the first column. \textbf{\textit{\#E}} denotes the number of EE-blocks. \textbf{\textit{Cost Percent of EE-Blocks}} shows the distribution of cost percent of EE-blocks. The given accuracy in the table is the testing accuracy.}
\label{table:cifar6eenet}
\end{table*}

\subsubsection{Comparison with previous work on CIFAR10}
It is not a trivial task to compare the performances of networks with adaptive computational structures. There are not any standard protocols, not every paper gives results on the same datasets, the concept of computational cost differs from work to work, the base networks are not always the same and source codes are not always available. Nevertheless, a number of studies present results on ImageNet \citep{imagenet} and Cifar10 \citep{cifar}. Due to our low budget for computational resources, we were not able to produce any EENet results on the ImageNet dataset\footnote{Our experiments are still running and we hope to include ImageNet results in a few weeks.} We collected CIFAR10 results from various papers in Table \ref{table:related_result}. Note that these results depend on the  implementation and training parameters (e.g. optimizer and learning rate), and also the base network used. To avoid the confusion, we have collected the results of ResNets and AlexNets \citep{alex} given in the these studies. Note that the number of layers of some of the studies are not specified since these are not given in the original papers.%The accuracy of our ResNets implementation is less than many of the studies. This may be due to our optimization configuration which we have employed them for both the ResNets and EENets. 
EENet with Pareto distribution yields a similar accuracy to that its counterpart ResNet with a relative cost of $50\%$. %s produce the close accuracy of our ResNet implementation (more accurate results can be obtained with smaller $\lambda$) by spending less computational cost than the given studies in Table \ref{table:related_result}. 

%In experiments, we have also observed that the additional parameters coming from the early-exit blocks are very small and can be ignored. For example, the number of parameters of an early-exit block of EENet-110 is $\sim$0.0002\% of the total parameters of the model with 10-classes datasets such as CIFAR10, MNIST and SVHN.

\begin{table*}
\centering
\begin{tabular}{|l||c|c|}
\hline
\multirow{2}{*}{\bf Model} & 
\multirow{2}{*}{\bf Accuracy} & 
\bf Relative \\
& & \bf Cost\\
\hline
\hline
ResNet-110  \citep{adanet}     & 94.39 & - \\
AdaNet-110  \citep{adanet}     & 94.24 & 0.82 \\
\hline
AlexNet \citep{branchynet}     & 78.38 & - \\
B-AlexNet \citep{branchynet}   & 79.19 & 0.42 \\
ResNet \citep{branchynet}      & 80.70 & - \\
B-ResNet \citep{branchynet}    & 79.17 & 0.53 \\
\hline
ResNet-110 \citep{skipnet}     & 93.60 & - \\
SkipNet-110 \citep{skipnet}    & 88.11 & 0.36 \\
\hline
Our ResNet-110         & 93.80 & - \\
EENet-110 (Pareto)     & 91.17 & 0.50 \\
EENet-110 (Fine)       & 85.93 & 0.36 \\
EENet-110 (GoldenRate) & 84.24 & 0.24 \\
EENet-110 (Linear)     & 87.94 & 0.38 \\
\hline
\multicolumn{3}{c}{}
\end{tabular}
\caption[Benchmark of related work on CIFAR10]{\textbf{Benchmark of related work on CIFAR10.} The computational cost rates are given in the columns of \textbf{\textit{Cost}}. The results are taken from the original papers. Note that results can change according to implementation details and training parameters (e.g. optimizer and learning rate). To avoid the confusion, we have shared the results of ResNets and AlexNets given in the these studies. Consequently, we can compare the results of only convenient work.}
\label{table:related_result}
\end{table*}

\section{Conclusion}
\label{section:b5}

In this paper, we propose the Early-exit Convolutional Neural Networks (EENets) which reduce the computational cost of convolutional neural networks (CNN) during inference. EENets have multiple exit blocks and they can classify input examples based on their characteristics at early stages of processing through these exit blocks. Thus, EENets terminate the execution early when possible and avoid wasting the computational cost on average. 

The early-exit (EE) blocks of EENets consist of a confidence branch and a classification branch. The confidence branch computes the confidence of the network in classification and exiting (i.e. stopping the inference process) at that location. On the other hand, the classification branch outputs a classification probability vector. Both branches are trainable and they are independent of each other.

EENets are trained with our proposed loss function which takes both the classical classification loss and the computational cost of inference into consideration. As a result,  confidence branches are adapted to the inputs so that less computation is spent for easy examples without harming  accuracy.% In other words, the confidence scores of the EE-blocks are trained in the accuracy and cost trade-off. Thus, they maintain the balance between them during inference.
Inference phase is similar to conventional feed-forward networks, however, when the output of a confidence branch reaches a constant threshold (i.e. $T = 0.5$), the inference stops for that specific example and it is classified by that exit block. 

\begin{sloppypar}
We conducted comprehensive experiments on MNIST, SVHN, CIFAR10 and Tiny-ImageNet datasets using both the 6n+2 and naive versions ResNets and their EENet counterparts. We  observed that EENets significantly reduce the computational cost (to 20\% of the original in ResNet-44 on CIFAR10) by maintaining the testing accuracy. 
\end{sloppypar}

Note that the idea behind EENets is applicable to any feed-forward neural network. However, in this paper, we demonstrated its use on convolutional neural networks and specifically on ResNets. Other types of networks (multi-layer perceptrons, recurrent neural networks) can be converted to their early-exit versions, too, which we leave as future work.

%\begin{acknowledgements}
%If you'd like to thank anyone, place your comments here
%and remove the percent signs.
%\end{acknowledgements}

% BibTeX users please use one of
\bibliographystyle{spbasic}
\bibliography{paper} % basic style, author-year citations
%\bibliographystyle{spmpsci}      % mathematics and physical sciences
%\bibliographystyle{spphys}       % APS-like style for physics
%\bibliography{}   % name your BibTeX data base

% Non-BibTeX users please use
%\nocite{resnet}

\end{document}